%% file: main.tex
\documentclass[10pt,twocolumn,letterpaper]{article}

\usepackage{cvpr}              
\input{preamble}
\definecolor{cvprblue}{rgb}{0.21,0.49,0.74}
\usepackage[pagebackref,breaklinks,colorlinks,allcolors=cvprblue]{hyperref}

\def\paperID{6160} 
\def\confName{CVPR}
\def\confYear{2026}

\title{Physically Interpretable Multi-Degradation Image Restoration via Deep Unfolding and Explainable Convolution}

\author{Hu Gao\\
Shanghai Jiao Tong University  \\
{\tt\small gao\_h@sjtu.edu.cn}
\and
Xiaoning Lei\\
CATL\\
{\tt\small leixn01@outlook.com}
\and
Xichen Xu\\
Shanghai Jiao Tong University  \\
{\tt\small neptune\_2333@sjtu.edu.cn}
\and
Depeng Dang\\
Beijing Normal University\\
{\tt\small ddepeng@mail.bnu.cn}
\and
Lizhuang Ma$^*$\\
Shanghai Jiao Tong University  \\
{\tt\small lzma@sjtu.edu.cn}
}

\begin{document}
\maketitle
\input{sec/0_abstract}    
\input{sec/1_intro}

\input{sec/2_relate}

\input{sec/3_method}

\input{sec/4_exper}
\input{sec/5_con}

{
    \small
    \bibliographystyle{ieeenat_fullname}
    \bibliography{main}
}

\input{sec/X_suppl}

\end{document}

%% file: sec/0_abstract.tex
\begin{abstract}
Although image restoration has advanced significantly, most existing methods target only a single type of degradation. In real-world scenarios, images often contain multiple degradations simultaneously, such as rain, noise, and haze, requiring models capable of handling diverse degradation types. Moreover, methods that improve performance through module stacking often suffer from limited interpretability. In this paper, we propose a novel interpretability-driven approach for multi-degradation image restoration, built upon a deep unfolding network that maps the iterative process of a mathematical optimization algorithm into a learnable network structure. Specifically, we employ an improved second-order semi-smooth Newton algorithm to ensure that each module maintains clear physical interpretability. To further enhance interpretability and adaptability, we design an explainable convolution module inspired by the human brain’s flexible information processing and the intrinsic characteristics of images, allowing the network to flexibly leverage learned knowledge and autonomously adjust parameters for different input. The resulting tightly integrated architecture, named InterIR, demonstrates excellent performance in multi-degradation restoration while remaining highly competitive on single-degradation tasks.
\end{abstract}

%% file: sec/1_intro.tex
\section{Introduction}
\label{sec:intro}
Image restoration (IR) aims to reconstruct a high-quality image $\overline{I}$ from its degraded observation $D$. The degradation process can be expressed as:
\begin{equation}
\label{eq:irf}
D = P \overline{I} + N
\end{equation}
where $P$ denotes the degradation operator and $N$ represents additive noise. 

Different choices of $P$ correspond to various image restoration tasks. In practice, however, the degradation process is often difficult to define precisely, leaving both $P$ and $N$ unknown. As a result, IR becomes a typical ill-posed problem, where multiple plausible solutions may correspond to the same degraded image. Traditional methods~\cite{2011Single, 10558778} tackle this issue by incorporating task-specific priors, constructing explicit degradation models, and performing inverse operations. Although these approaches are physically interpretable and sometimes effective, they depend heavily on strong assumptions about degradation factors such as noise and blur kernels. In real-world conditions, however, the degradation is typically uncertain and complex, making accurate modeling challenging and limiting the generalization ability of such methods.

In recent years, the field of image restoration has experienced a paradigm shift driven by advances in deep learning~\cite{PGH2Netisu2025prior,FSNet, efderainguo2025efficientderain+}. By leveraging the statistical characteristics of natural images, deep learning-based methods can implicitly learn diverse priors, achieving significantly better performance than traditional approaches. However, these data-driven networks, typically built through module stacking, often suffer from excessive parameter redundancy, limited interpretability. To improve model interpretability, some image restoration methods~\cite{deepun10.1145/3664647.3681532,deepun10487002,deepun10803115,deepun10830558,deepunDutta2022DIVADU} based on deep unfolding networks map the iterative solutions of mathematical optimization algorithms into learnable network structures, fully integrating model priors. This significantly enhances both the interpretability of network inference and the robustness of network learning. However, most deep unfolding-based methods are tailored to specific tasks or impose strict constraints on the degradation process, making it challenging to design a robust network capable of handling multiple restoration tasks under complex conditions. Some approaches~\cite{DeepSN-Net10820096, VLUNetZeng_2025_CVPR} attempt to address this by reformulating the original nonlinear coupled system into a network-friendly convex optimization problem.

While these methods~\cite{DeepSN-Net10820096, VLUNetZeng_2025_CVPR} allow a single model to handle various degradation types, they are mainly focused on single-degradation image restoration (SDIR) and typically assume that each image contains only one type of degradation. In real-world scenarios, however, images are often heterogeneous, dynamic, and uncertain, with multiple degradations such as rain, haze, and noise occurring simultaneously. This complexity limits the effectiveness of current approaches in multi-degradation restoration. Some works~\cite{AASO,FDTANetgao2025frequency,tanet,OWN,DIGNet} have investigated multi-degradation image restoration (MDIR), but their performance remains restricted, partly because they rely solely on simple attention mechanisms to select relevant features. Ref-IRT~\cite{REF-IRT} attempts to overcome this by using a multi-stage framework that progressively transfers similar edges and textures from reference images. Nevertheless, it still depends heavily on the model’s representational power and does not address MDIR from an interpretability standpoint.

\begin{figure} 
    \centerline{\includegraphics[width=1\linewidth]{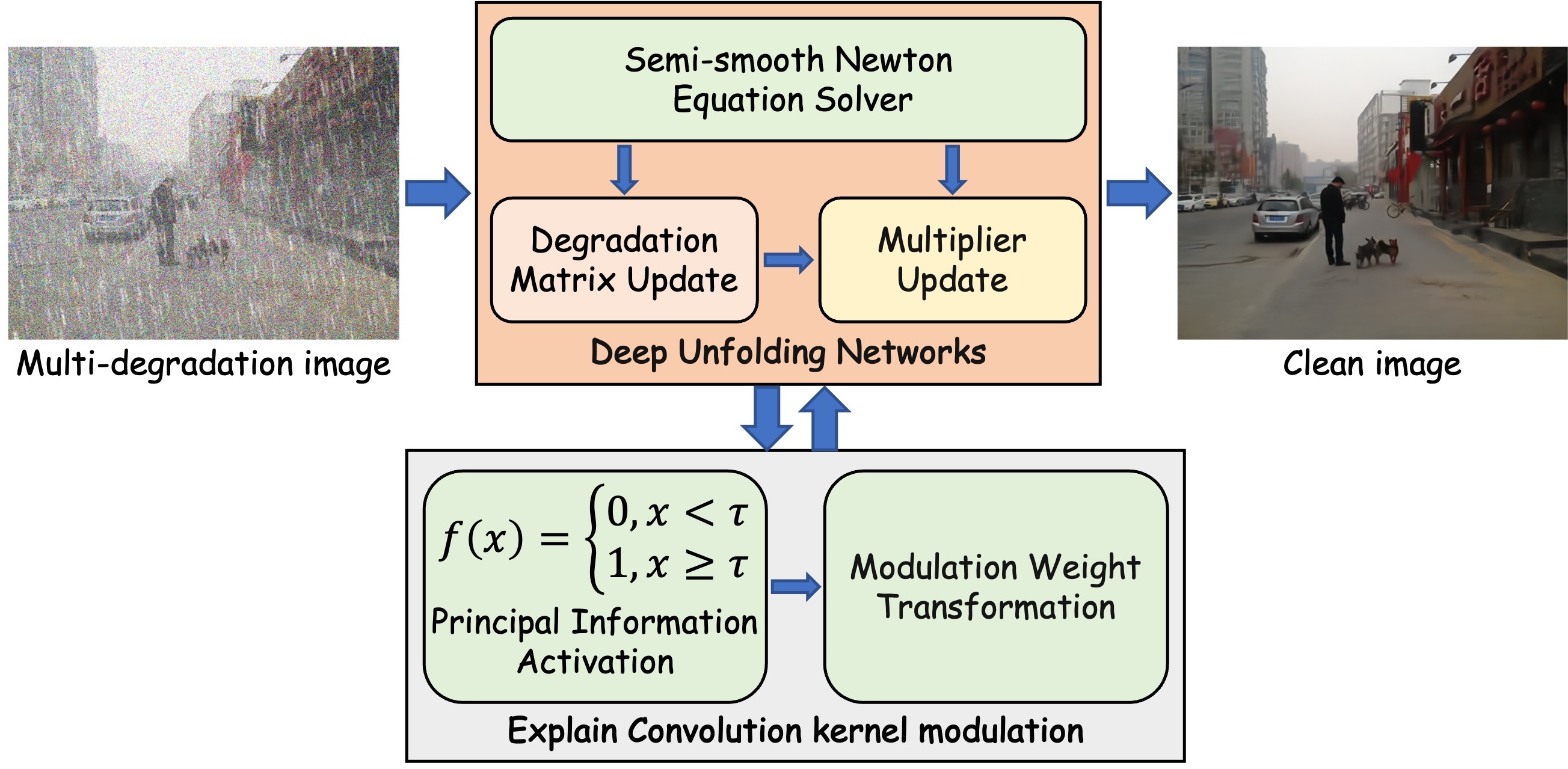}}
	\caption{Mechanisms of our method.}
 \label{fig:exam}
\end{figure}

Given these considerations, a natural question arises: can we design a network that adaptively restores images with multiple degradations by explicitly representing and dynamically regulating its internal decision-making process? To address this challenge, we propose InterIR (Figure~\ref{fig:exam}), an adaptive MDIR network by strictly following the improved semi-smooth Newton algorithm in its modules and connections, providing strong physical interpretability and transparency, while integrating an explainable convolution module to achieve image-specific adaptive modeling and iterative parameter optimization. Specifically, we reformulate the image restoration task as an improved second-order semi-smooth Newton optimization problem. The parameters of this optimization are learned through the network, ensuring that each module operates with clear physical interpretability.
Furthermore, inspired by the human brain’s flexible information processing and the intrinsic characteristics of images, we design an explainable convolution module. This module enables principal information activation and modulation weight transformation, allowing the network to dynamically adjust convolution kernel parameters based on multi-degradation image features, thereby achieving image-specific adaptive modeling. Extensive experiments show that our method delivers outstanding performance on multi-degradation restoration and remains highly competitive on single-degradation tasks.

The main contributions of this work are:
\begin{enumerate}
	\item We propose InterIR for MDIR, which follows the improved semi-smooth Newton algorithm in its modules and connections for strong physical interpretability and transparency, and incorporates an explainable convolution module for image-specific adaptive modeling and iterative parameter optimization.
    \item We reformulate the image restoration task as an improved second-order semi-smooth Newton optimization problem, with parameters learned through the network to ensure that each module operates with clear physical interpretability.
	\item We design an explainable convolution module that flexibly leverages learned knowledge and autonomously adjusts parameters based on multi-degradation image features, enabling image-specific adaptive modeling.
    \item Extensive experiments demonstrate that our method achieves outstanding performance in MDIR, effectively handling complex combinations of degradations, while also maintaining strong competitiveness on SDIR.
\end{enumerate}

%% file: sec/2_relate.tex
\section{Related Works}
\label{sec:relate}

\subsection{Image Restoration}
SDIR aims to recover high-quality images affected by a specific type of corruption. Traditional methods~\cite{2011Single, 10558778} tackle this ill-posed problem by introducing handcrafted priors to constrain the solution space. While effective in certain scenarios, these priors are highly dependent on expert knowledge and often lack adaptability.

With the rapid advancement of deep learning, numerous data-driven approaches~\cite{LSSRgao2024learning,adarevD10656920,FSNet,aclgu2025acl,VLUNetZeng_2025_CVPR} have been developed to enhance image restoration performance. LSSR~\cite{LSSRgao2024learning} introduces an adaptive module combination strategy for lightweight stereo image super-resolution, while XYScanNet~\cite{liu2024xyscannet} employs an alternating intra- and inter-slice scanning strategy to better capture spatial dependencies. NAFNet~\cite{chen2022simple} streamlines the model by refining baseline components and removing or replacing non-linear activations. SFNet~\cite{SFNet} and FSNet~\cite{FSNet} propose dynamic frequency selection modules that adaptively identify the most informative components for restoration. MambaIR~\cite{guo2025mambair} introduces a four-directional unfolding strategy with channel attention to enhance spatial representation, and MambaIRV2~\cite{guo2025mambairv2} further integrates non-causal modeling for improved expressiveness. AdaRevD~\cite{AdaRevD} employs a classifier to estimate the degradation level of image patches, though it lacks adaptive flexibility. LoFormer~\cite{xintm2024LoFormer} utilizes local channel-wise self-attention in the frequency domain to capture cross-covariance patterns within both low- and high-frequency local regions.

However, the above methods generally train a single model dedicated to one specific degradation type, making it difficult to construct a unified framework capable of simultaneously addressing multiple degradation scenarios within a single network. CAPTNet~\cite{CAPTNet10526271} leverages changes in data components through prompt learning to achieve integrated image restoration. IDR~\cite{IDRzhang2023ingredient} introduces an ingredient-oriented perspective to model degradation components, enhancing the framework’s scalability across diverse conditions. AdaIR~\cite{cui2025adair} disentangles degradation information from clean image content by jointly exploiting both spatial and frequency representations.  Despite their promising results in SDIR, these methods encounter significant challenges in MDIR, where complex, intertwined degradations severely hinder restoration accuracy and generalization.


MDIR aims to recover a clean image from one affected by multiple types of degradation. AASO~\cite{AASO} executes parallel restoration operations with attention-based weighting to select the optimal strategy, but feature heterogeneity limits its performance on multi-degraded images. OWAN~\cite{OWN} fuses heterogeneous features via high-order tensors to exploit richer statistics, yet its complex design hampers fine-detail reconstruction.
DIGNet~\cite{DIGNet} models sequential and spatially varying degradations, while MEPSNet~\cite{Kim_2020_ACCV} adopts a mixture-of-experts framework with parameter sharing to handle region-specific distortions.

Despite these advancements, their performance remains constrained, as most focus on task-adaptive attention designs that execute multiple parallel operations weighted by attention to select the most suitable restoration strategy. To address this, Ref-IRT~\cite{REF-IRT} adopts a three-stage framework: the first stage estimates residuals in a coarse-to-fine manner, while the subsequent stages progressively transfer fine details from reference images. FDTANet~\cite{FDTANetgao2025frequency} performs adaptive restoration in the frequency domain by analyzing component differences. Although some approaches~\cite{DefusionLuo_2025_CVPR, Perceive-IR10990319} are primarily tailored for SDIR, they also demonstrate strong MDIR performance through enhanced representational power from large vision models. However, these methods still overlook the intricate interactions among different degradation mechanisms. In this work, we address this limitation by modeling from the perspective of degradation mechanisms, mapping the iterative process of a mathematical optimization algorithm into a learnable network structure. Based on image-specific characteristics, we realize MDIR through an interpretable and adaptive framework.

\subsection{Deep Unfolding Networks}
Deep unfolding network is a neural architecture that integrates model-based optimization and data-driven learning by unfolding the iterative steps of a traditional mathematical optimization algorithm into a sequence of learnable network layers. Each layer corresponds to one iteration of the optimization process, where the algorithmic parameters are learned from data, enabling the network to retain physical interpretability. 
DIVA~\cite{deepunDutta2022DIVADU} unfolds a baseline adaptive denoising algorithm within a deep neural network framework, drawing inspiration from quantum many-body physics.
D$^3$U-Net~\cite{deepun10.1145/3664647.3681532} introduces a dual-domain collaborative optimization strategy that fuses visual representations from the image domain with multi-resolution information from the wavelet domain.
PEUNet~\cite{deepun10830558} incorporates a snow shape prior to generate supervision signals, addressing the lack of real snow mask annotations.
VLU-Net~\cite{VLUNetZeng_2025_CVPR} leverages vision-language model features to automatically extract degradation-aware cues, eliminating the need for manually defined categories.
REPO~\cite{deepun10487002} proposes a high-accuracy rotation-equivariant proximal network that embeds rotation symmetry priors into the deep unfolding framework.
DeepSN-Net~\cite{DeepSN-Net10820096} reformulates an improved second-order semi-smooth Newton (ISN) algorithm, transforming the original nonlinear equations into an optimization problem amenable to network implementation.
Although the above methods assign physical meaning to each iteration, they still rely on neural networks to update matrices and solve operators, leaving the feature transformation and decision-making process largely a “black box.” To address this limitation, we introduce an explainable convolution module inspired by the human brain’s flexible information processing and the intrinsic properties of images, enabling the network to adaptively utilize learned knowledge and autonomously adjust parameters for different inputs.

%% file: sec/3_method.tex
\section{Method}
\label{sec:method}
In this section, we first present an overview of the InterIR framework, followed by the formulation of the restoration problem and a detailed description of the solving process. Finally, we introduce the explainable convolution module.

\begin{figure*} 
    \centerline{\includegraphics[width=1\linewidth]{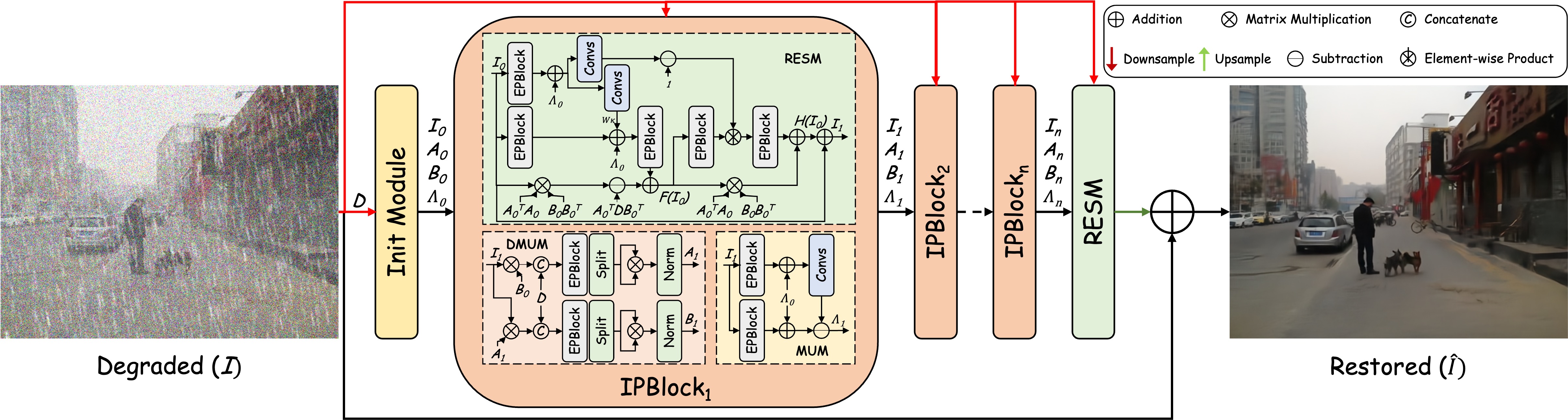}}
	\caption{The overall architecture of the proposed InterIR consists of an initial module followed by $n$ interpretability blocks (IPBlocks). Each IPBlock contains a restoration equation solver module (RESM), a degradation matrix update module (DMUM), and a multiplier update module (MUM).}
 \label{fig:network}
\end{figure*}

\subsection{Overall Pipeline} 
Our proposed InterIR, illustrated in Figure~\ref{fig:network}. 
Given a degraded image $\mathbf{D} \in \mathbb{R}^{H \times W \times 3}$, InterIR first apply $4 \times $ downsampling on it to obtain a downsampled tensor $\mathbf{F_{0}} \in \mathbb{R}^{H_d \times W_d \times C_d}$ ($H_d$, $W_d$, and $C_d$ denote the height, width, and number of channels, respectively). The downsampled tensor then processed through a init module to initialize degradation matrix, multiplier and restored image. These initialized variables then go
through $n$ interpretability blocks (IPBlocks) and a restoration equation solver module (RESM) to obtain the final  residual image $\mathbf{I_R} \in \mathbb{R}^{H \times W \times 3}$, which is added to the degraded input to produce the restored image: $\mathbf{\hat{I}} = \mathbf{I_R} + \mathbf{D}$.

To enable effective restoration in both the spatial and frequency domains, we optimize the proposed InterIR network using the following loss function:
\begin{equation}
\begin{aligned}
\label{eq:loss1}
L &= L_{1}(\hat{I},\overline I) + \lambda |\mathcal{F}(\hat{I}) - \mathcal{F}(\overline I)|_1
\end{aligned}
\end{equation}
where $\overline I$ denotes the target images, and $L_1$ is the L1 loss.  $\mathcal{F}$ denoting the fast Fourier transform. $\lambda$ balances the contributions of the two loss terms and is set to 0.1, following~\citep{FSNet}.

\subsection{Problem Statement} 
The image degradation process is typically formulated as Eq.~\ref{eq:irf}. However, the degradation matrix $P$ grows quadratically with spatial resolution, leading to substantial computational overhead. To alleviate this issue, inspired by LoRA~\cite{hu2022lora}, we decompose $P$ into two smaller degradation matrices via a Kronecker product, effectively reducing computational complexity and allowing the degradation operation to scale linearly with image resolution. Specifically, we factorize $P \in \mathbb{R}^{H \times W \times C}$ into $A \in \mathbb{R}^{H \times H \times C}$ and $B \in \mathbb{R}^{W \times W \times C}$, enabling Eq.~\ref{eq:irf} to be reformulated as
\begin{equation}
\label{eq:irf2}
D = A  \overline{I}  B + N
\end{equation}
where $B^{T} \otimes A = P$. Following~\cite{DeepSN-Net10820096}, to enable adaptive restoration under complex and diverse real-world degradations, we impose multiple constraints on the model and define the following optimization objective:
\begin{equation}
\label{eq:444}
\min_{\overline{I}, A, B} \ | A \overline{I} B - D |_2^2 + \alpha g(V(\overline{I})) + \beta \phi(A) + \gamma \varphi(B)
\end{equation}
where $g(\cdot)$, $\phi(\cdot)$, and $\varphi(\cdot)$ are regularization functions, and $\alpha$, $\beta$, and $\gamma$ denote the corresponding weight coefficients. The operator $V(\cdot)$ extracts shallow image features.
To facilitate optimization, we introduce an auxiliary variable $C$ to substitute $V(\overline{I})$ and reformulate the problem using the augmented Lagrangian method as follows:
\begin{equation} 
\begin{aligned} 
\label{eq:aim} 
\mathcal{L}_{I, A, B, V(\overline{I}), \Lambda} &:= \frac{1}{2} \| A I B - D \|_2^2 + \alpha g(C) + \beta \phi(A) + \gamma \varphi(B) 
\\ &\quad + \langle \Lambda, V(\overline{I}) - C \rangle + \frac{\epsilon}{2} \| V(\overline{I}) - C \|_2^2 
\end{aligned} 
\end{equation}
where $\Lambda$ is the multiplier and $\epsilon$ is the penalty coefficient.

\subsection{Solving Process}
To solve the optimization objective in Eq.~\ref{eq:aim}, we adopt the improved second-order semi-smooth Newton (ISN) algorithm~\cite{DeepSN-Net10820096}, which decomposes the problem into three subproblems to alternately update $(\overline{I}_n, C_n)$, $(A_n, B_n)$, and $\Lambda_n$:
\begin{equation}
\begin{aligned}
\label{eq:6c}
(\overline{I}_n, C_{n}) &= \arg\min_{\overline{I}, C} \mathcal{L}(\overline{I}, C, A_{n-1}, B_{n-1}, \Lambda_{n-1}) 
\\
(A_n, B_n) &= \arg\min_{A, B} \mathcal{L}(\overline{I}_n, C_{n}, A, B, \Lambda_{n-1})  
\\
\Lambda_n &= \Lambda_{n-1} + \epsilon \left( V(\overline{I}_n) - C_n \right)
\end{aligned}
\end{equation}

We now focus on solving the the core of the ISN algorithm in Eq.~\ref{eq:6c}.
Given the updated parameters $\Lambda = \Lambda_{n-1}$, $A = A_{n-1}$, and $B = B_{n-1}$, the optimal solution can be obtained by setting:
\begin{equation}
\begin{aligned}
\label{eq:7b}
\alpha \, \partial g(C) - \Lambda + \epsilon (C - V(\overline{I})) &\ni 0 
\\
A^T (A \overline{I} B - D) B^T + V^T [ \Lambda + \epsilon (V(\overline{I}) - C) ] &= 0
\\
\end{aligned}
\end{equation}
where $\partial g(C)$ denotes the subderivative of $g(C)$.
Then, it can be inferred that:
\begin{equation}
\begin{aligned}
\label{eq:7bc}
\\ C = \frac{\Lambda}{\sigma} + V(\overline{I}) - \frac{\partial g(C)}{\sigma} = S(\frac{\Lambda}{\sigma} + V(\overline{I}))
\\
A^T (A \overline{I} B - D) B^T + V^T [ \Lambda + \epsilon (V(\overline{I}) - S(\frac{\Lambda}{\sigma} + V(\overline{I}))) ] &= 0
\end{aligned}
\end{equation}
where $S(\cdot)$ is a nonlinear function determined by $\partial g(\cdot)$ and the ratio $\alpha / \sigma$. For simplicity, we denote the resulting function as $F(\overline{I})$ and solve it iteratively using gradient descent:
\begin{equation}
\begin{aligned}
F(\overline{I}_{n-1}) &= A^T (A \overline{I}_{n-1} B - D) B^T \\
&+ V^T [ \Lambda + \epsilon (V(\overline{I}_{n-1}) - S(\frac{\Lambda}{\sigma} + V(\overline{I}_{n-1})))] 
\\
H(\overline{I}_{n-1}) &= A^T A F(\overline{I}_{n-1}) B B^T 
\\&
+ \sigma V^T[ V(F(\overline{I}_{n-1}) \odot (\mathbf{1} - S_d(\frac{\Lambda}{\sigma} + V(\overline{I}_{n-1})))]
\\
\overline{I}_{n} &\gets \overline{I}_{n-1} - \eta H(\overline{I}_{n-1}) 
 \label{eq:1oo}
\end{aligned}
\end{equation}
where $H(\overline{I}_{n-1}) \in \partial \| F(\overline{I}_{n-1}) \|_2^2$ is the subderivative with respect to $\overline{I}_{n-1}$, $S_d(\cdot)$ is the subderivative of $S(\cdot)$, and $\eta$ is the step size. The solution of the  sub-problem can be obtained as:
\begin{equation}
\begin{aligned}
 A_n &= D_\phi(D, A_{n-1} \overline{I}_n), B_n = D_\varphi(D, \overline{I}_n A_n)
\\
\Lambda_n &= \Lambda_{n-1} + \epsilon [V(\overline{I}_{n}) - S(\frac{\Lambda_{n-1}}{\sigma} + V(\overline{I}_{n}))]
 \label{eq:10c}
\end{aligned}
\end{equation}
where $D_\phi(\cdot)$ and $D_\varphi(\cdot)$ are determined by the regularization functions $\phi(\cdot)$ and $\varphi(\cdot)$, respectively.

The above procedure follows the derivation presented in DeepSN-Net~\cite{DeepSN-Net10820096} to introduce the solving process of the ISN algorithm. Based on Eq.~\ref{eq:10c}, we design a corresponding network structure that ensures each module maintains strong physical interpretability. As illustrated in Figure~\ref{fig:network}, given a degraded image $\mathbf{D}$, the network first downsamples it and initializes the degradation matrices, multipliers, and restored image parameters as:
\begin{equation}
\begin{aligned}
\label{eq:init}
I_0& = D_{\downarrow_4}, \Lambda_0 = f_1(I_0) + I_0
\\
A_0 &= \frac{I_0I_0^T}{||I_0I_0^T||_2} 
,B_0 = \frac{I_0^TI_0}{||I_0^TI_0||_2} 
\end{aligned}
\end{equation}

These initialized variables are subsequently fed through $n$ interpretability blocks (IPBlocks). Each IPBlock comprises a restoration equation solver module (RESM) constructed according to Eq.~\ref{eq:1oo}, a degradation matrix update module (DMUM), and a multiplier update module (MUM) following Eq.~\ref{eq:10c}. In particular, given the initial estimate $I_0$, the RESM computes $I_1$ as follows:
\begin{equation}
\begin{aligned}
\label{eq:fi0}
F(I_0) =& A_0^T(D \ominus A_0I_0B_0)B_0^T 
\\
& \oplus EP(EP(I_0) \oplus \Lambda_0 \oplus W_\kappa\text{Convs}(\Lambda_0 \oplus EP(I_0)))
\\
H(I_0) =&  A_0^TA_0F(I_0)B_0B_0^T 
\\
&\oplus EP(1\ominus \text{Convs} (\Lambda_0 \oplus EP(I_0))) 
\textcircled{\textasteriskcentered} EP(F(I_0))
\\
I_1 =& I_0 \otimes - \eta H(I_0)
\end{aligned}
\end{equation}
where $EP(\cdot)$ denotes the explainable block (EPBlock),  $W_\kappa$ is a learnable weight, and $\text{Convs}$ consists of multiple convolutions followed by a ReLU. Subsequently, the DMUM is employed to update the degradation matrix as follows:
\begin{equation}
\begin{aligned}
\label{eq:dm}
A^0_1, A^1_1 &= Split(EP(Concat(I_1 \otimes B_0,D))) 
\\
A_1 &= Norm(A^0_1\otimes A^1_1)
\\
B^0_1, B^1_1 &= Split(EP(Concat(A_1 \otimes I_1,D))) 
\\
B_1 &= Norm(B^0_1\otimes B^1_1)
\end{aligned}
\end{equation}

Finally, we use the MUM to update the multiplier as:
\begin{equation}
\label{eq:mul}
\Lambda_1 = \Lambda_0 \oplus EP(I_1) \ominus Convs(\Lambda_0 \oplus EP(I_1))
\end{equation}

It is worth noting that, unlike DeepSU-Net~\cite{DeepSN-Net10820096}, in Eq.~\ref{eq:fi0} we do not explicitly solve 
$F(\cdot)$ and $H(\cdot)$ as in Eq.~\ref{eq:1oo} \textbf{(More details are provided in the Supplementary material)}. Instead, we introduce additional networks within both the RESM and DMUM to learn the weighting parameters between variables, making our formulation more consistent with Eq.~\ref{eq:444}.
Ablation studies further confirm that this design leads to improved performance.

\begin{figure} 
    \centerline{\includegraphics[width=1\linewidth]{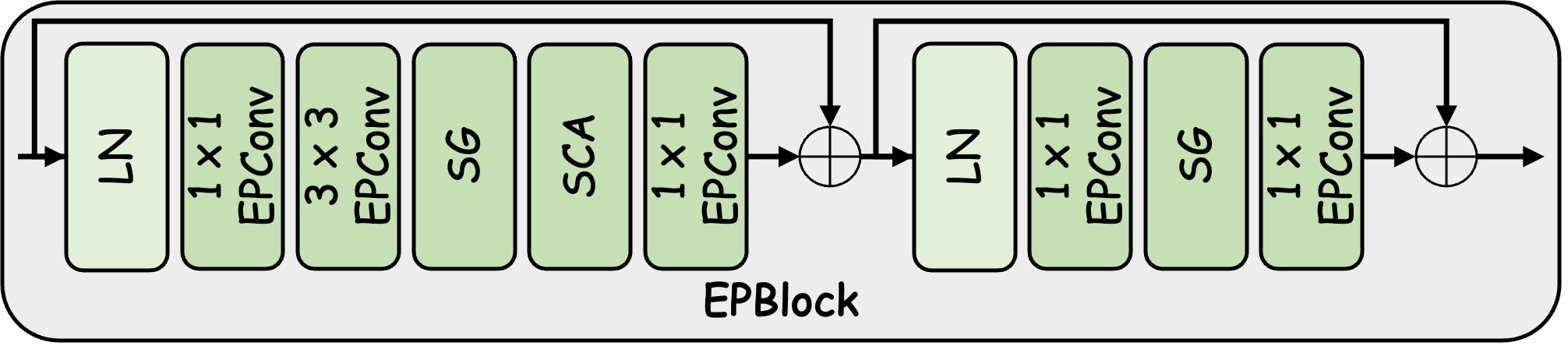}}
	\caption{The structure of explainable  block (EPBlock).}
 \label{fig:epb}
\end{figure}

\subsection{Explainable  Block} 
While the operations above provide interpretability for each restoration step, we find that relying solely on the constraints in Eq.~\ref{eq:444} is insufficient for fully adaptive multi-degradation restoration. 
To enhance both interpretability and input adaptability in convolutional operations, we design an explainable block (EPBlock), which not only improves MDIR performance but also mitigates the black-box limitations of DeepSU-Net~\cite{DeepSN-Net10820096} in learning matrices and multiplier, yielding a more interpretable restoration framework. 

As shown in Figure~\ref{fig:epb}, the EPBlock follows the structure of NAFBlock~\cite{chen2022simple}, with the key distinction that we replace the standard convolution with an explainable convolution. Inspired by principles from brain science and visual perception, this convolution combines the human brain's flexible information processing with the intrinsic characteristics of images. It dynamically modulates the convolution kernel for each input sample without modifying the original weights. Unlike conventional convolutions that use a fixed kernel for all samples, our design introduces an input-specific attention mask to adaptively adjust the effective receptive weights according to the input content, enabling more transparent and adaptive multi-degradation image restoration.
Given an input feature map $x \in \mathbb{R}^{B \times C \times H \times W}$, a standard convolution computes  as:
\begin{equation}
y = \text{Conv}(x) = W * x + \text{bias},
\end{equation}
where 
$W \in \mathbb{R}^{C_{\text{out}} \times C_{\text{in}} \times K_h \times K_w}$  is the  kernel.
In contrast, the explainable convolution introduces an input-specific modulation. For each input sample, we first generate a binary attention mask $M$ based on a learnable threshold $\tau$ to highlight the most informative features, helping the model identify the types of multiple degradations in the image:
\begin{equation}
M_{b,c,h,w} =
\begin{cases}
1, & x_{b,c,h,w} \ge \tau, \\
0, & \text{otherwise}.
\end{cases}
\end{equation}

\begin{table*}
\centering
\caption{Quantitative results on various combinations of degradation, where H, R, and N denote haze, rain, and noise, respectively.}
\label{tb:mix}
\resizebox{\linewidth}{!}{
\begin{tabular}{cccccccc||c}
    \hline
   Methods& H + R + N  & H + R &H + N& R + N & H &R&N&Average
    \\
    \hline\hline
    Restormer~\cite{Zamir2021Restormer} &23.52/0.792 &25.45/0.836 &26.73/0.863	&25.31/0.815	&29.82/0.931	&28.05/0.873	&27.44/0.837 &26.62/0.850
     \\
    AirNet~\cite{all_conli2022all} &27.41/0.812 &28.28/0.889 &27.59/0.861 &26.98/0.821 &30.22/0.959 &28.27/0.881 & 27.25/0.847 &28.00/0.867
       \\
    U$^2$Former~\cite{u2former} &25.07/0.803 &26.23/0.856 &26.79/0.864	&26.02/0.816	&29.95/0.933	&28.50/0.876	&27.12/0.831 &27.09/0.854
     \\
    PromptIR~\cite{potlapalli2023promptir} &27.54/0.819	&28.43/0.901 &27.99/0.871	&27.05/0.822	&30.46/0.956	&28.78/0.885	&27.92/0.851 &28.31/0.872
       \\
    VLU-Net~\cite{VLUNetZeng_2025_CVPR} &29.35/0.842	&30.38/0.935 &28.79/0.869	&28.21/0.843	&31.37/0.959	&33.82/0.968	&27.88/0.853 &29.97/0.896
       \\
FDTANet~\cite{FDTANetgao2025frequency}  &29.68/0.846	&30.91/0.937	&\textbf{29.06}/\underline{0.874}	&27.63/0.827	&\textbf{31.91/0.981} &29.13/0.892	&28.88/0.857 &29.60/0.888
\\
DeepSN-Net~\cite{DeepSN-Net10820096} & 29.11/0.841 & 30.01/0.921 & 28.43/0.864 & 27.99/0.842 & 31.42/0.959 & 33.82/0.969 & 28.01/0.852 &29.83/0.893
\\
Ref-IRT~\cite{REF-IRT}&28.95/0.823	&\underline{32.33/0.961} &28.62/0.872	&31.22/0.831	&30.69/0.961	&31.89/0.902	&29.39/0.899 &30.44/0.893
\\
Perceive-IR~\cite{Perceive-IR10990319}&\textbf{30.22/0.894}	&31.55/0.945	&28.32/0.871	&30.52/0.849	&\underline{31.90/0.980}	&38.95/0.983	&31.88/0.923	&31.90/0.921
\\
Defusion~\cite{DefusionLuo_2025_CVPR}&29.66/0.849	&29.93/0.937	&28.11/0.866	&\textbf{32.17/0.920}	&31.29/0.969	&\textbf{40.15/0.986}	&\underline{32.72/0.925}	&\underline{32.01/0.922}
\\
\hline
\textbf{InterIR(Ours)} &\textbf{30.22}/\underline{0.880}	&\textbf{32.46/0.963}			&\textbf{29.06/0.883}	&\underline{31.85/0.885}	&31.88/0.968	&\underline{39.33/0.984}	&\textbf{32.75/0.926} &\textbf{32.51/0.927}
    \\ 
    \hline
\end{tabular}}
\end{table*}

\begin{figure*} 
    \centerline{\includegraphics[width=1\linewidth]{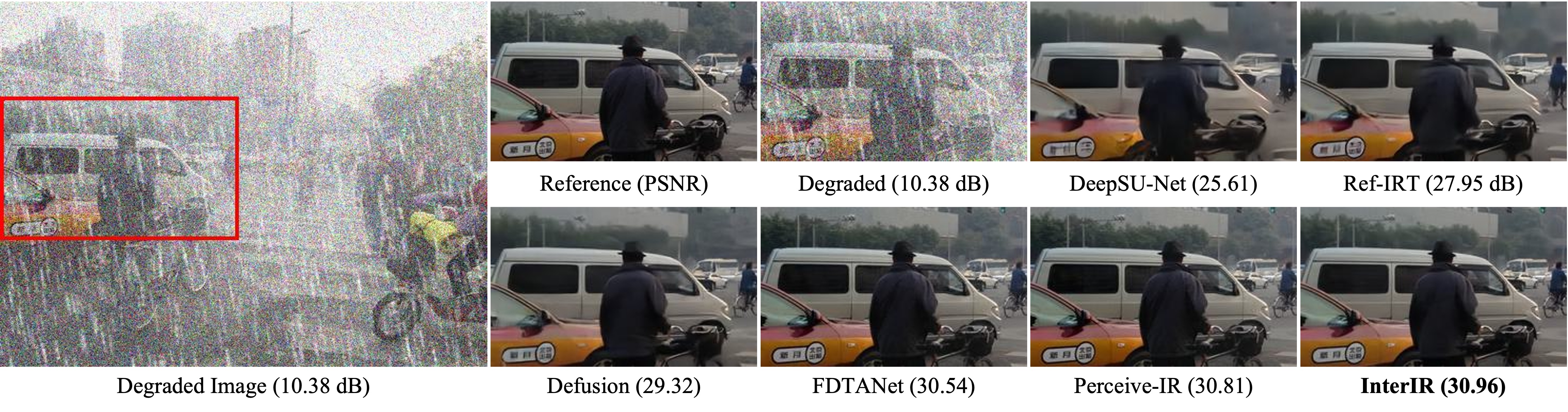}}
	\caption{Qualitative results under the MDIR experimental setup. Our InterIR produces images that are visually closer to the ground truth.}
 \label{fig:mix}
\end{figure*}

To align with the convolution kernel size, the mask is spatially pooled:
\begin{equation}
\tilde{M}_{b,c} = \text{Pool}(M_{b,c}) \in \mathbb{R}^{K_h \times K_w}.
\end{equation}

Next, a soft attention distribution is obtained via softmax normalization:
\begin{equation}
A_{b,c,i,j} = \frac{\exp(\tilde{M}_{b,c,i,j})}{\sum_{i',j'} \exp(\tilde{M}_{b,c,i',j'})}, 
\quad \text{s.t. } \sum_{i,j} A_{b,c,i,j} = 1.
\end{equation}

The attention map $A$ acts as a modulation factor, producing an adaptive kernel for each input:
\begin{equation}
W_b = W \odot A_b,
\end{equation}
where $\odot$ denotes element-wise multiplication and $W_b$ is the dynamically modulated kernel for the $b$-th input.
Finally, the output is computed via batch-independent convolution:
\begin{equation}
y_b = W_b * x_b + \text{bias}, \quad b = 1, 2, \ldots, B.
\end{equation}

This design offers several advantages. First, each input sample generates its own kernel modulation $A_b$, enabling the network to adaptively focus on the most relevant spatial regions. Second, the learned mask $M$ explicitly highlights which regions activate the convolution kernel, providing an interpretable view of the model's attention. Third, since $W_b$ is produced through modulation rather than replacing $W$, the original kernel remains trainable and shared across the dataset, ensuring stability during learning.

%% file: sec/4_exper.tex
\begin{figure*} 
    \centerline{\includegraphics[width=1\linewidth]{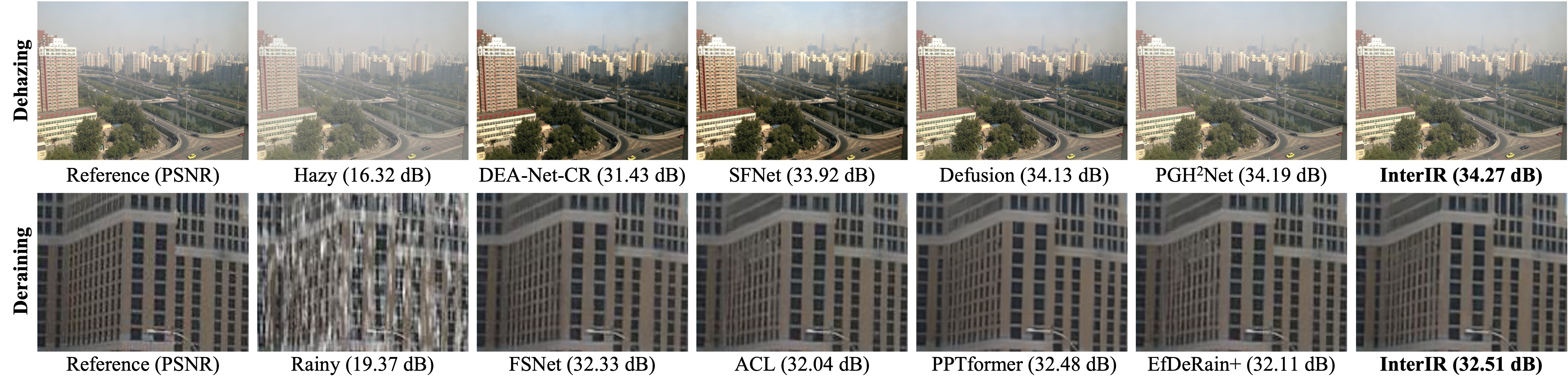}}
	\caption{Qualitative results under the SDIR experimental setup. Our InterIR is able to reconstruct finer and sharper details.}
 \label{fig:sig}
\end{figure*}

\begin{table}
    \centering
        \caption{Image dehazing results.}
    \label{tab:sot}
    \resizebox{\linewidth}{!}{
    \begin{tabular}{c|cccc}
    \hline
    \multicolumn{1}{c|}{} & \multicolumn{2}{c}{SOTS-Indoor}  & \multicolumn{2}{c}{SOTS-Outdoor} 
    \\
   Methods & PSNR $\uparrow$ & SSIM $\uparrow$  & PSNR $\uparrow$ & SSIM $\uparrow$ 
   \\
   \hline
   \hline
IRNext~\cite{IRNeXt}&41.21 &\textbf{0.996} &39.18 &\textbf{0.996}       
\\
DEA-Net-CR~\cite{deanetchen2024dea} & 41.31&\underline{0.995} & 36.59 & 0.990 
\\
 Defusion~\cite{DefusionLuo_2025_CVPR} & 41.65&\underline{0.995} & 37.41 & \underline{0.993}
\\
PGH$^2$Net~\cite{PGH2Netisu2025prior}&\underline{41.70} &\textbf{0.996}&37.52 &0.989
         \\
         \hline
        \textbf{ InterIR(Ours)} &\textbf{41.98}	&\textbf{0.996}	&\textbf{39.77}	&0.989
         \\
         \hline
    \end{tabular}}
\end{table}

\begin{table*}
\centering
\caption{Image deraining results.}
\label{tb:derain}
    \resizebox{\linewidth}{!}{
\begin{tabular}{c|cccccccc||cc}
    \hline
    \multicolumn{1}{c|}{} & \multicolumn{2}{c}{Test100~\citep{Test100}}  & \multicolumn{2}{c}{Test1200~\citep{MSPFN}} & \multicolumn{2}{c}{Rain100H~\citep{Rain100}} & \multicolumn{2}{c||}{Rain100L~\citep{Rain100}} & \multicolumn{2}{c}{Average} 
    \\
   Methods &PSNR $\uparrow$ &  SSIM $\uparrow$  & PSNR $\uparrow$ & SSIM $\uparrow$ &PSNR $\uparrow$ &SSIM $\uparrow$ & PSNR $\uparrow$&SSIM $\uparrow$ &PSNR $\uparrow$ & SSIM $\uparrow$
    \\
    \hline\hline
    FSNet~\citep{FSNet} &31.05&0.919 &33.08&0.916 & 31.77&0.906 &38.00 & 0.972 &33.48 &0.928
    \\
    MHNet~\citep{gao2025mixed} &31.25 &0.901 &\underline{33.45} &0.925 &31.08 &0.899 &\underline{40.04} &\textbf{0.985} &33.96 &0.928
     \\
     PPTformer~\cite{pptformerwang2025intra} & 31.48 & \textbf{0.922} & 33.39 & 0.911 &  31.77 & 0.907
     & 39.33 & \underline{0.983} & 33.99 & 0.931
     \\
 ACL~\cite{aclgu2025acl} &31.51 & 0.914 & 33.27 &0.928  & \underline{32.22} & \underline{0.920} & 39.18 &\underline{0.983}  & 34.05 & \underline{0.936}
     \\
DeepSN-Net~\cite{DeepSN-Net10820096} &\underline{31.60} & \underline{0.920} &\underline{33.45} &\underline{0.931} & 31.81 & 0.904 & 38.59 & 0.975 &33.86 & 0.933
     \\
     EfDeRain+~\cite{efderainguo2025efficientderain+} &31.10 &0.911 &33.12 & 0.925 & \textbf{34.57} &\textbf{0.957} & 39.03 & 0.972 &\underline{34.46} & \textbf{0.941}
     \\
      \hline
      \textbf{InterIR(Ours)}  & \textbf{32.65}	&0.919	&\textbf{35.02}	&\textbf{0.940}		&31.50	&0.898	&\textbf{40.83}	&\textbf{0.985}	&\textbf{35.00}	&\underline{0.936}
    \\
    \hline
\end{tabular}}
\end{table*}

\section{Experiments}
\label{sec:exp}
In this section, we first describe the experimental setup, followed by qualitative and quantitative comparison results. Finally, we present ablation studies to validate the effectiveness of our approach. \textbf{Due to page limits, more experiments we show in the supplementary material.}

\subsection{Experimental Setup}
We conducted experiments under both MDIR and SDIR settings.
\textbf{Datasets.} For the MDIR setting, we adopt the dataset introduced by~\cite{FDTANetgao2025frequency}. In the SDIR setting, the model is trained on images collected from multiple datasets~\cite{Rain100,Test100,8099669,7780668}, and evaluated on several benchmark test sets, including Rain100H~\cite{Rain100}, Rain100L~\cite{Rain100}, Test100~\cite{Test100}, and Test1200~\cite{DIDMDN} for the image deraining task. For image dehazing, training and evaluation are conducted on the RESIDE dataset~\citep{RESIDEli2018benchmarking}, with testing performed on its SOTS~\citep{RESIDEli2018benchmarking}.

\textbf{Training details.} Our models are optimized using Adam~\cite{2014Adam} with $\beta_1 = 0.9$ and $\beta_2 = 0.999$. The learning rate is initialized at $2 \times 10^{-4}$ and gradually reduced to $1 \times 10^{-7}$ according to the cosine annealing strategy~\cite{2016SGDR}. During training, image patches of size $256 \times 256$ are randomly sampled with a batch size of 32 for a total of $4 \times 10^5$ iterations. Data augmentation is performed through horizontal and vertical flipping. The $n$ in Figure~\ref{fig:network} is 16. We set the $\eta = 0.01$  in Eq.~\ref{eq:fi0}. To ensure fairness, all compared deep learning-based methods are either fine-tuned or retrained following the parameter configurations specified in their original papers.

\subsection{Experimental Results}

\subsubsection{MDIR Setting}
We evaluate InterIR on seven different combinations of degradation types, covering all permutations of haze, rain, and noise. Table~\ref{tb:mix} presents the quantitative comparison results. Averaged across all tasks, our method outperforms the second-best approach, Defusion~\cite{DefusionLuo_2025_CVPR}, by 0.5 dB. Compared with other MDIR methods, Ref-IRT~\cite{REF-IRT} and FDTANet~\cite{FDTANetgao2025frequency}, our model achieves significant improvements of 2.07 dB and 2.91 dB, respectively. Compared with the baseline model DeepSN-Net~\cite{DeepSN-Net10820096}, our InterIR achieves consistently better performance across all evaluations.

To further assess the robustness of our approach across diverse degradation types, we conduct additional experiments using the same training dataset. The results indicate that while InterIR may not always produce the largest gain in every specific scenario, it consistently delivers stable performance across an arbitrary number of degradation combinations.
As illustrated in Figure~\ref{fig:mix}, our model generates restored images that are sharper and visually closer to the ground truth compared to competing methods.

\subsubsection{SDIR Setting}
To demonstrate that the proposed InterIR is not only effective for MDIR but also exhibits strong generalization to SDIR tasks, we evaluate it on two representative scenarios: image dehazing and image deraining. As shown in Figure~\ref{fig:sig}, InterIR produces visually superior results, significantly mitigating color distortion and restoring natural tones compared with state-of-the-art methods. Furthermore, our model reconstructs sharper textures and finer structural details, highlighting its robust restoration capability.

\textbf{Image Dehazing.} Table~\ref{tab:sot} reports the quantitative results of different image dehazing methods. Overall, InterIR consistently achieves superior performance across both indoor and outdoor scenarios, demonstrating its effectiveness and robustness. Specifically, on the indoor dataset SOT-Indoor, InterIR surpasses the previous best-performing method, PGH$^2$Net~\cite{PGH2Netisu2025prior}, by 0.28 dB in PSNR. On the outdoor dataset SOT-Outdoor, it further achieves a 0.59 dB improvement over the previous state-of-the-art method, IRNext~\cite{IRNeXt}.

\textbf{Image Deraining.} Following the evaluation protocol in prior work~\cite{gao2025mixed}, we report PSNR and SSIM metrics on the Y channel of the YCbCr color space for the image deraining task. As presented in Table~\ref{tb:derain}, InterIR consistently matches or surpasses existing methods across all four benchmark datasets. Notably, it achieves an average PSNR improvement of 0.54 dB over the previous best-performing method, EfDeRain+~\cite{efderainguo2025efficientderain+}, and demonstrates a substantial 1.14 dB gain over our baseline DeepSN-Net~\cite{DeepSN-Net10820096}, underscoring its strong deraining performance and robustness.

\begin{table}
    \centering
    \caption{Ablation study on individual components of InterIR.}
    \label{tab:abl1}
       \resizebox{\linewidth}{!}{
    \begin{tabular}{ccc}
    \hline
         Method&  PSNR &$\triangle$ PSNR 
         \\
         \hline
         Baseline & 29.11 & -
         \\
         replace with Explain Convolution& 29.83 & +0.72
        \\
         replace with  RESM& 29.44&  +0.33 
        \\
        replace with DMUM and MUM & 29.35& +0.24
         \\
         Ours & 30.22& +1.11
         \\
         \hline
    \end{tabular}}
\end{table}

\subsection{Ablation Studies}
We perform ablation studies in the MDIR setting to evaluate the effectiveness of our proposed modules. Using DeepSN-Net~\cite{DeepSN-Net10820096} as the baseline, we conduct a stepwise ablation by progressively integrating each module.

\subsubsection{Effectiveness of Each Module}
Table~\ref{tab:abl1} presents the results of ablation experiments evaluating the contribution of each component in InterIR. The baseline network achieves a PSNR of 29.11 dB. Replacing the standard convolution with the proposed explainable convolution module raises the PSNR to 29.83 dB, a gain of 0.72 dB, demonstrating its effectiveness in adaptively leveraging image-specific features for restoration. Adding the RESM module increases the PSNR to 29.44 dB, a 0.33 dB improvement over the baseline, while integrating the DMUM and MUM modules yields a PSNR of 29.35 dB, corresponding to a 0.24 dB gain. Finally, the complete InterIR model, incorporating all proposed modules, achieves the highest PSNR of 30.22 dB, an overall improvement of 1.11 dB. These results indicate that each module contributes positively to the network’s performance, and their combination produces the most significant enhancement in multi-degradation image restoration.

\begin{figure} 
    \centerline{\includegraphics[width=1\linewidth]{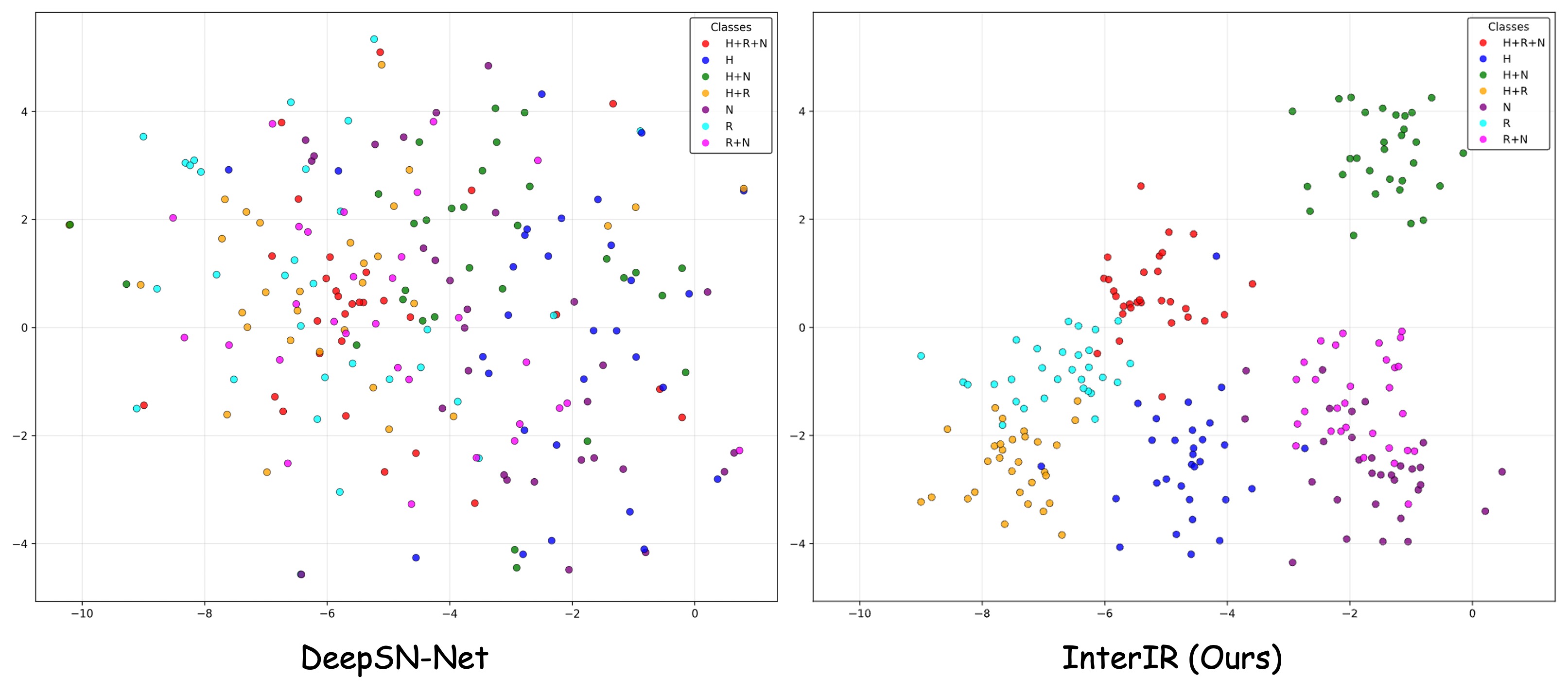}}
	\caption{The figure presents t-SNE visualizations of degradation embeddings from InterIR (ours) and DeepSN-Net~\cite{DeepSN-Net10820096}. }
 \label{fig:tsne}
\end{figure}

To further validate the effectiveness of our design, Figure~\ref{fig:tsne} presents t-SNE visualizations of the degradation embeddings from InterIR (ours) and the baseline model DeepSN-Net~\cite{DeepSN-Net10820096}. Different colors correspond to different degradation types. In our model, the embeddings for each task are more distinctly clustered, demonstrating the ability of image-specific adaptive modeling to capture discriminative degradation contexts that facilitate restoration. In contrast, the baseline model fails to clearly separate different degradation types, which aligns with the unsatisfactory MDIR performance reported in Table~\ref{tb:mix}.

\begin{table}
    \centering
    \caption{Plug-and-play ablation experiments.}
    \label{tab:ablplug}
    \begin{tabular}{ccc}
    \hline
         Method&  PSNR &$\triangle$ PSNR 
         \\
         \hline
         FDTANet~\cite{FDTANetgao2025frequency}& 29.68 & -
         \\
         replace with Explain Convolution& 30.31 & +0.63
        \\
        \hline
         DeepSN-Net~\cite{DeepSN-Net10820096}& 29.11 & - 
        \\
        replace with Explain Convolution & 29.83& +0.72     
         \\
         \hline
    \end{tabular}
\end{table}

\subsubsection{Generality of the  Explainable Convolution}
To further assess the generality and effectiveness of the explainable convolution module, we integrate it into the convolution layers of existing image restoration methods. As reported in Table~\ref{tab:ablplug}, this modification results in substantial performance improvements. Specifically, FDTANet~\cite{FDTANetgao2025frequency} and DeepSN-Net~\cite{DeepSN-Net10820096} achieve PSNR gains of 0.63 dB and 0.72 dB, respectively, demonstrating that our explainable convolution can consistently enhance both existing architectures and their restoration capabilities.

%% file: sec/5_con.tex
\section{Conclusion}
\label{sec:con}
In this paper, we present InterIR, an interpretability-driven framework for multi-degradation image restoration.  Based on a deep unfolding architecture, our approach transforms the iterative process of a second-order semi-smooth Newton optimization algorithm into a learnable and physically interpretable network structure. In addition, the inclusion of an explainable convolution module,  allows InterIR to dynamically adjust its parameters according to the characteristics of each input image, thereby enhancing both interpretability and adaptability. Extensive experiments show that InterIR achieves SOTA performance in MDIR tasks and remains highly competitive in SDIR scenarios.

%% file: sec/X_suppl.tex
\clearpage
\setcounter{page}{1}
\maketitlesupplementary

\section{Overview}
\label{sec:over}
The Appendix is composed of:

Dataset~\ref{sec:data}

DeepSU-Net~\cite{DeepSN-Net10820096} vs. InterIR~\ref{sec:deevsin}

More Experiments~\ref{sec:mas}

Additional Visual Results~\ref{sec:Visual}

\section{Dataset}
\label{sec:data}

\subsection{MDIR}
Following~\cite{FDTANetgao2025frequency}, we perform comparative experiments on a dataset that includes various combinations of degradations such as haze, rain, and noise. The dataset contains 18,000 natural images collected from multiple public image restoration benchmarks~\citep{RESIDEli2018benchmarking,Rain100,Test100,BSDmartin2001database}. Among them, 13,000 images are randomly selected for training, where synthetic degradations are applied with intensity ranges of [0,150] for haze, [0,300] for rain, and [0,50] for noise. From the remaining images, 500 are randomly chosen for testing. As summarized in Table~\ref{tab:testco}, the test set includes seven distinct degradation combinations, covering all possible permutations of haze, rain, and noise. For instance, in the “haze + rain + noise” scenario, all three degradation types are applied as in training, whereas in the “haze + rain” case, only the noise component is omitted while the other settings remain unchanged. The remaining configurations are constructed following the same principle.

\subsection{SDIR}
\subsubsection{Image Deraining}
Following the experimental settings of recent state-of-the-art image deraining methods~\cite{Zamir2021MPRNet,FSNet}, we train our model on 13,712 clean–rain image pairs sourced from multiple benchmarks~\cite{Rain100,Test100,8099669,7780668}. The trained InterIR is then evaluated on four widely used test sets: Rain100H~\cite{Rain100}, Rain100L~\cite{Rain100}, Test100~\cite{Test100}, and Test1200~\cite{DIDMDN}.

\subsubsection{Image Dehazing}
To evaluate the image dehazing performance, we conduct experiments on the daytime subset of the RESIDE dataset~\cite{RESIDEli2018benchmarking}, which includes both synthetic indoor and outdoor scenes. RESIDE provides two training sets: the indoor training set (ITS) and the outdoor training set (OTS), as well as a synthetic objective testing set (SOTS). The ITS contains 13,990 hazy images generated from 1,399 clean images, while the OTS includes 313,950 hazy images synthesized from 8,970 clean images. We train our model separately on the ITS and OTS datasets and evaluate it on their corresponding test sets, SOTS-Indoor and SOTS-Outdoor, each containing 500 paired samples.

\begin{table}
    \centering
    \caption{Test set generation by multi-degradation types.}
    \label{tab:testco}
    \resizebox{\linewidth}{!}{
    \begin{tabular}{cccc}
        \hline
      Multi-degradation & Haze level & Rain level & Noise level
      \\
      \hline
      Haze + Rain + Noise & [0,150] & [0,300] & [0,50]
      \\
       Haze + Rain  & [0,150] & [0,300] & [0]
       \\
       Haze  + Noise & [0,150] & [0] & [0,50]
        \\
          Rain  + Noise & [0] & [0,300] & [0,50]
        \\
          Haze  & [0,150] & [0] & [0]
        \\
          Rain & [0] & [0,300] & [0]
        \\
         Noise & [0] & [0] & [0,50]
        \\
        \hline
    \end{tabular}} 
\end{table}

\section{DeepSU-Net~\cite{DeepSN-Net10820096} vs. InterIR}
\label{sec:deevsin}
In this work, we employ the improved second-order semi-smooth Newton (ISN) algorithm proposed in DeepSU-Net~\cite{DeepSN-Net10820096} to preserve clear physical interpretability for each module. 

First, we all decomposes the image restoration problem into three subproblems to alternately update $(\overline{I}_n, C_n)$, $(A_n, B_n)$, and $\Lambda_n$:
\begin{equation}
\begin{aligned}
\label{eq:supp6c}
(\overline{I}_n, C_{n}) &= \arg\min_{\overline{I}, C} \mathcal{L}(\overline{I}, C, A_{n-1}, B_{n-1}, \Lambda_{n-1}) 
\\
(A_n, B_n) &= \arg\min_{A, B} \mathcal{L}(\overline{I}_n, C_{n}, A, B, \Lambda_{n-1})  
\\
\Lambda_n &= \Lambda_{n-1} + \epsilon \left( V(\overline{I}_n) - C_n \right)
\end{aligned}
\end{equation}

Then, we all use the gradient descent algorithm to get the solution of  with the following iterative steps:
\begin{equation}
\begin{aligned}
F(\overline{I}_{n-1}) = &A^T (A \overline{I}_{n-1} B - D) B^T 
\\
&+ V^T [ \Lambda + \epsilon (V(\overline{I}_{n-1}) - S(\frac{\Lambda}{\sigma} + V(\overline{I}_{n-1})))] 
\label{eq:su10c1}
\end{aligned}
\end{equation}

\begin{equation}
\begin{aligned}
H(\overline{I}_{n-1}) =& A^T A F(\overline{I}_{n-1}) B B^T 
\\
&+ \sigma V^T[ V(F(\overline{I}_{n-1}) \odot (\mathbf{1} - S_d(\frac{\Lambda}{\sigma} + V(\overline{I}_{n-1}))))]
\label{eq:su10c2}
\end{aligned}
\end{equation}

\begin{equation}
\overline{I}_{n} \gets \overline{I}_{n-1} - \eta H(\overline{I}_{n-1})
\label{eq:su10c3}
\end{equation}

\begin{equation}
A_n = D_\phi(D, A_{n-1} \overline{I}_n)
\label{eq:su10c4}
\end{equation}

\begin{equation}
B_n = D_\varphi(D, \overline{I}_n A_n)
\label{eq:su10c41}
\end{equation}

\begin{equation}
\Lambda_n = \Lambda_{n-1} + \epsilon [V(\overline{I}_{n}) - S(\frac{\Lambda_{n-1}}{\sigma} + V(\overline{I}_{n}))]
\label{eq:su10c5}
\end{equation}

Finally, the transformed convex optimization problem was further unfolded into a learnable network structure. In DeepSU-Net~\cite{DeepSN-Net10820096}, this process was implemented as follows:

\begin{equation}
\begin{aligned}
&F(\overline{I}_{n-1}) = A^T (A \overline{I}_{n-1} B - D) B^T 
\\
&+ f_E(\Lambda_{n-1} + f_E(\overline{I}_{n-1}) - f_S(\Lambda_{n-1} + f_E(\overline{I}_{n-1})))
\label{eq:su11c1}
\end{aligned}
\end{equation}

\begin{equation}
\begin{aligned}
&H(\overline{I}_{n-1}) = A^T A F(\overline{I}_{n-1}) B B^T 
\\
&+ f_E[f_E(F(\overline{I}_{n-1})) \odot (1-f_S(\Lambda_{n-1} + f_E(\overline{I}_{n-1})))]
\label{eq:su11c2}
\end{aligned}
\end{equation}

\begin{equation}
\overline{I}_{n} \gets \overline{I}_{n-1} - \eta H(\overline{I}_{n-1})
\label{eq:su11c3}
\end{equation}

In our work, we do not expand it into this form, but instead:
\begin{equation}
\begin{aligned}
&F(\overline{I}_{n-1}) = A^T (A \overline{I}_{n-1} B - D) B^T 
\\
&+ f_E[\Lambda_{n-1} + f_E(\overline{I}_{n-1}) + W_\kappa f_S(\Lambda_{n-1} + f_E(\overline{I}_{n-1}))]
\label{eq:su12c1}
\end{aligned}
\end{equation}

\begin{equation}
\begin{aligned}
&H(\overline{I}_{n-1}) = A^T A F(\overline{I}_{n-1}) B B^T 
\\
&+ f_E[(1 - f_S(\Lambda_{n-1} + f_E(F(\overline{I}_{n-1}) ) \odot  f_E(F(\overline{I}_{n-1}))]
\label{eq:su12c2}
\end{aligned}
\end{equation}

\begin{equation}
\overline{I}_{n} \gets \overline{I}_{n-1} - \eta H(\overline{I}_{n-1})
\label{eq:su12c3}
\end{equation}

From the above, it is evident that the main difference is Eq.~\ref{eq:su11c1} and Eq.~\ref{eq:su12c1}. In our approach, we do not compute $F(\cdot)$ using the subtraction form in Eq.~\ref{eq:su10c1}. Instead, we use an additive form and introduce an additional weight-learning module in the model to control the computation through a learnable weight matrix. The primary reason is that during training, gradient vanishing often occurs. Our investigation revealed that when the depth expansion follows the form of Eq.~\ref{eq:su11c1}, the result can become zero, leading to training collapse during backpropagation. To address this, we replaced the subtraction with addition and introduced learnable weights $W_\kappa $ through an additional network. This modification facilitates training, better preserves feature information, and allows flexible tuning of feature contributions.

\section{More Experiments}
\label{sec:mas}

\subsection{Alternative of Explain Convolution}
To further assess the effectiveness of the proposed explain convolution, we substitute it with two representative convolutional variants, CoMo~\cite{dcm10256095} and ARConv~\cite{ARC11093597}, and conduct comparative experiments under identical settings. As summarized in Table~\ref{tab:alhfs}, replacing our explain convolution with either alternative results in a noticeable performance drop. Specifically, the model using CoMo decreases the PSNR by 0.38 dB, while the ARConv-based model suffers a reduction of 0.29 dB. Moreover, both alternatives lead to a significant increase in model parameters, with ARConv introducing an additional 10.0M parameters. These results clearly demonstrate that our explain convolution achieves a better balance between restoration accuracy and model compactness, validating its design advantage in effectively extracting image-specific features while maintaining computational efficiency.

\begin{table}
    \centering
       \caption{Alternative of Explain Convolution.}
    \label{tab:alhfs}
    \begin{tabular}{cccc}
    \hline
         Net& PSNR & $\triangle$ PSNR &Params(M)
         \\
         \hline
          Explain Convolution & 32.51 & - & -
         \\
            CoMo~\cite{dcm10256095} & 32.13 & - 0.38 & +0.5
            \\
            ARConv~\cite{ARC11093597} & 32.22 & - 0.29 & +10.0
            \\
           
         \hline
    \end{tabular}
\end{table}

\begin{table}
    \centering
    \caption{The evaluation of model computational complexity. }
    \label{tab:computational}
     \resizebox{\linewidth}{!}{
    \begin{tabular}{cccccc}
    \hline
         Method& Time(s) & Params(M) &Flops(G) & PSNR & SSIM
         \\
         \hline\hline
         VLU-Net~\cite{VLUNetZeng_2025_CVPR} &0.743 &35 & 143 &29.97 & 0.896
         \\
         PromptIR~\cite{potlapalli2023promptir} &1.012 & 33& 134& 28.31 & 0.872
         \\
         Perceive-IR~\cite{Perceive-IR10990319} &0.682 &45 &144&31.90 & 0.921 
         \\
        DeepSU-Net~\cite{DeepSN-Net10820096} &0.331 &24 & 86 &29.83 &0.893
         \\
         \hline
        \textbf{InterIR(Ours)} &0.334 &27 &22 &32.51 & 0.927
        \\
         \hline
    \end{tabular}}
\end{table}

\subsection{Resource Efficient}
We further evaluate the computational efficiency of InterIR by comparing its runtime, parameter count, and FLOPs with recent state-of-the-art methods. As shown in Table~\ref{tab:computational}, InterIR not only delivers superior restoration performance but also substantially reduces computational cost. Specifically, it surpasses the previous best method, Perceive-IR~\cite{Perceive-IR10990319}, by 0.61 dB while using only 50.2\% of its inference time. Compared with the baseline model DeepSU-Net~\cite{DeepSN-Net10820096}, although our method introduces additional modules for learning weighting parameters—resulting in slightly higher inference time and parameter count—it avoids a multi-scale architecture, leading to a significant reduction in FLOPs.

\begin{figure*} 
    \centerline{\includegraphics[width=1\textwidth]{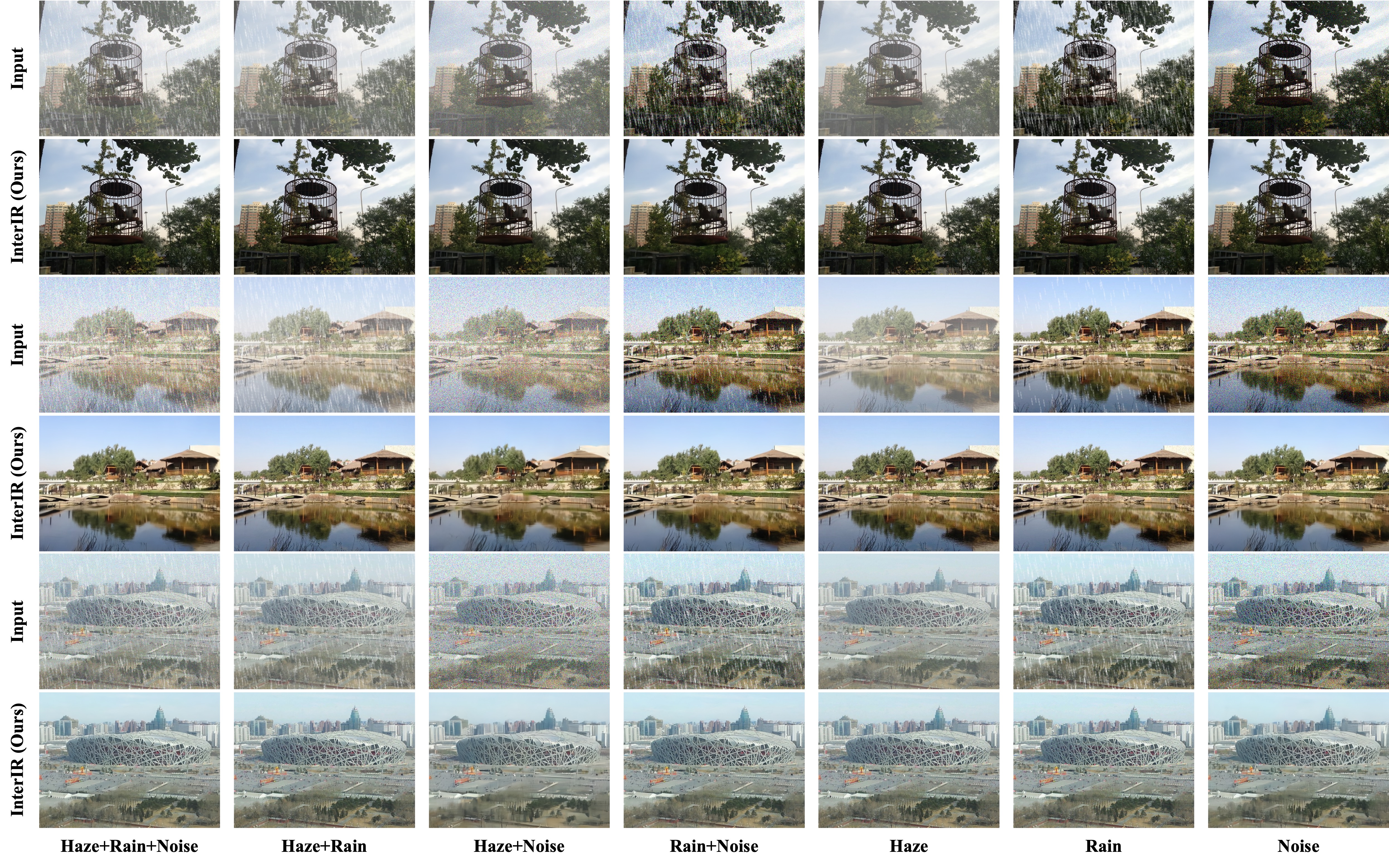}}
	\caption{Image restoration results across various combinations of degradation types. The top row displays the degraded inputs,  and the bottom row shows the corresponding restoration results generated by InterIR.}
 \label{mix_v}
\end{figure*}

\section{Additional Visual Results}
\label{sec:Visual}

Figure~\ref{mix_v} presents the visual results of our method under different combinations of degradation types. For each case, the top row shows the input images suffering from multiple degradations, while the bottom row displays the corresponding outputs restored by InterIR. It can be observed that our model effectively eliminates diverse degradations and consistently reconstructs images with sharp details, natural textures, and high visual fidelity across all scenarios.